\documentclass[sigconf]{aamas}

\usepackage{balance}
\usepackage{algorithm}
\usepackage{algpseudocode}
\usepackage{tabularx}
\usepackage{makecell}
\usepackage{subcaption}
\usepackage{multirow}
\usepackage{pdfpages}

\doi{EXQH7884}

\makeatletter
\gdef\@copyrightpermission{
  \begin{minipage}{0.2\columnwidth}
   \href{https://creativecommons.org/licenses/by/4.0/}{\includegraphics[width=0.90\textwidth]{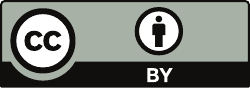}}
  \end{minipage}\hfill
  \begin{minipage}{0.8\columnwidth}
   \href{https://creativecommons.org/licenses/by/4.0/}{This work is licensed under a Creative Commons Attribution International 4.0 License.}
  \end{minipage}
  \vspace{5pt}
}
\makeatother

\setcopyright{ifaamas}
\acmConference[AAMAS '26]{Proc.\@ of the 25th International Conference
on Autonomous Agents and Multiagent Systems (AAMAS 2026)}{May 25 -- 29, 2026}
{Paphos, Cyprus}{C.~Amato, L.~Dennis, V.~Mascardi, J.~Thangarajah (eds.)}
\copyrightyear{2026}
\acmYear{2026}
\acmDOI{}
\acmPrice{}
\acmISBN{}

\title[AAMAS-2026 Formatting Instructions]{SOM: Structured Opponent Modeling for LLM-based Agents \\via Structural Causal Model}

\author{Shiyue Cao}
\affiliation{
  \institution{School of Artificial Intelligence, University of Chinese Academy of Sciences \&  Institute of Automation, Chinese Academy of Sciences}
  \city{Beijing}
  \country{China}
}
\email{caoshiyue2021@ia.ac.cn}

\author{Pei Xu}
\authornote{Pei Xu and Kaiqi Huang are corresponding authors.}
\affiliation{
  \institution{National Key Laboratory of Cognition and Decision Intelligence for Complex Systems, Institute of Automation, Chinese Academy of Sciences}
  \city{Beijing}
  \country{China}
}
\email{pei.xu@ia.ac.cn}

\author{Likun Yang}
\affiliation{
  \institution{School of Artificial Intelligence, University of Chinese Academy of Sciences}
  \city{Beijing}
  \country{China}
}
\email{yanglikun2021@ia.ac.cn}

\author{Lei Cui}
\affiliation{
  \institution{School of Artificial Intelligence, University of Chinese Academy of Sciences }
  \city{Beijing}
  \country{China}
}
\email{cuilei2024@ia.ac.cn}

\author{Xiaotang Chen}
\affiliation{
  \institution{Institute of Automation, Chinese Academy of Sciences}
  \city{Beijing}
  \country{China}
}
\email{xtchen@nlpr.ia.ac.cn}

\author{Kaiqi Huang}
\authornotemark[1]
\affiliation{
  \institution{School of Artificial Intelligence, University of Chinese Academy of Sciences \&  Institute of Automation, Chinese Academy of Sciences}
  \city{Beijing}
  \country{China}
}
\email{kaiqi.huang@nlpr.ia.ac.cn}

\begin{abstract}

Accurately predicting opponents' behavior from interactions is a fundamental capability for large language model (LLM)-based agents in multi-agent and game-theoretic environments. Existing approaches often entangle opponent modeling with prediction, relying on implicit contextual reasoning and limiting adaptability in dynamic interactions. To this end, 
we propose \textbf{S}tructured  \textbf{O}pponent \textbf{M}odeling (\textbf{SOM}), a two-stage opponent modeling framework that distinctly decouples opponent model construction and opponent prediction.
At the construction stage, SOM employs a Structural Causal Model (SCM), a graph-based formalism for representing dependencies among variables, to capture directed links between opponents' observations and actions, yielding an explicit and structured opponent representation.
At the prediction stage, the LLM performs structured reasoning along clear pathways derived from the SCM, improving both prediction accuracy and stability. 
Extensive experiments on diverse multi-agent benchmarks demonstrate that SOM consistently outperforms state-of-the-art LLM-based reasoning baselines, enabling more accurate and adaptable strategic decision-making in complex and dynamic multi-agent interactions.
\vspace{-2.0em}
\end{abstract}


\newcommand{\BibTeX}{\rm B\kern-.05em{\sc i\kern-.025em b}\kern-.08em\TeX}

\begin{document}

\pagestyle{fancy}
\fancyhead{}

\maketitle

\section{Introduction}

Large Language Models (LLMs) have emerged as a transformative development in artificial intelligence. By training on vast amounts of text data, they acquire extensive world knowledge \cite{sun2023head} and exhibit strong reasoning \cite{imani2023mathprompter} and problem-solving \cite{rasal2024llm} abilities. These powerful capabilities have positioned LLMs as promising candidates for autonomous agents in complex, interactive environments such as economic simulations \cite{horton2023large, li2024econagent}, collaborative tasks \cite{chen2024comm}, and strategic negotiations \cite{bianchi2024well}. In these multi-agent settings, an agent's success critically hinges on its ability to model opponent behavior and adapt its own strategy accordingly \cite{nashed2022survey}, and a lack of deep awareness of the opponent's behavior can lead to strategies that are easily exploited or misaligned, resulting in suboptimal outcomes \cite{carroll2019utility}. This is particularly crucial in strategic reasoning scenarios characterized by complex strategic interactions and continuously evolving behaviors.

However, current approaches tend to implicitly entangle the modeling---the process of identifying how opponents make decisions\allowbreak ---with opponent prediction through LLM-based contextual reasoning~\cite{zhang2024proagent,xu2023exploring,guan2024richelieu,guo2023suspicion}. This approach lacks a clear, controllable reasoning path---it neither specifies how to systematically establish the link between raw observations and an opponent's final action, nor does it guide the language model on what key intermediate reasoning processes to include, such as inferring the opponent's beliefs or their hidden information.  Without this structural guidance, the language model's inference process becomes difficult to control, often missing key information~\cite{liu2023lost} or producing hallucinations~\cite{ji2023survey}. 
While existing structured reasoning methods such as Tree-of-Thought~\cite{yao2023tree} and Graph-of-Thought \cite{Besta2024GraphofThoughtsSE} enhance LLM reasoning in many tasks, they are primarily designed for static problem settings and lack mechanisms to incorporate external feedback, making them difficult to adapt to the non-stationary nature of strategic interactions~\cite{zhang2024llm}.
These limitations highlight the need for new approaches that enable explicit and adaptable opponent modeling in dynamic multi-agent settings. 

\begin{figure}[t]
    \centering
    \includegraphics[width=\linewidth]{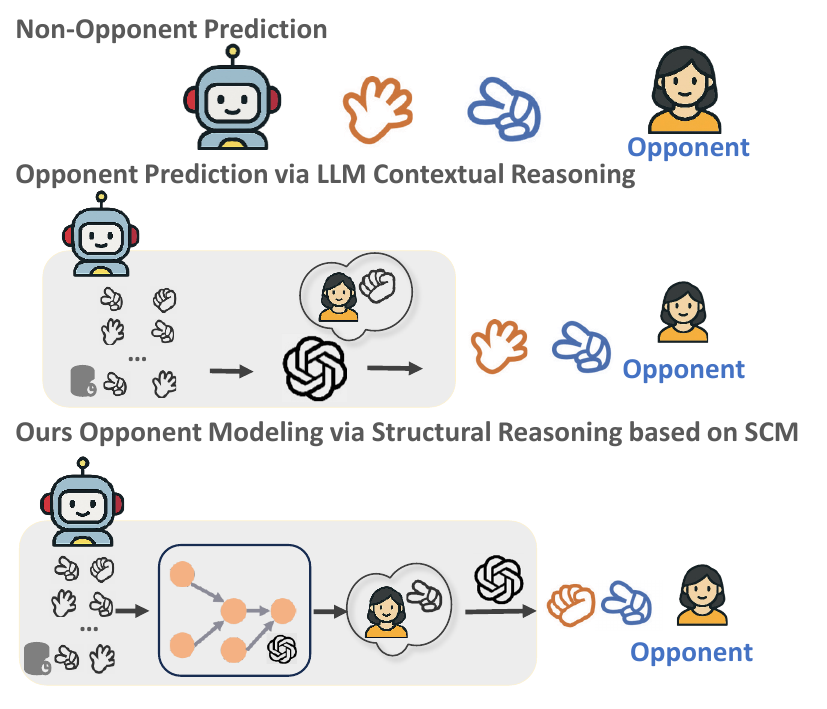}\\[ -1pt]

    \caption{ Illustrating different opponent modeling paradigms. Unlike baselines that ignore opponent behavior or entangle modeling within implicit reasoning, SOM explicitly constructs a structured model to guide opponent prediction.}
    \label{fig:opp}
    \vspace*{-2em}
\end{figure}

To address these challenges, we propose \textbf{S}tructured \textbf{O}pponent \textbf{M}odeling (\textbf{SOM}), a two-stage framework that explicitly separates opponent model construction and opponent prediction. This design enables LLM-based agents to reason about opponents through a structured and controllable process rather than relying solely on implicit contextual inference.
As illustrated in Figure~\ref{fig:opp}, this two-stage design offers explicit and controllable reasoning pathways, in contrast to existing LLM-based approaches that entangle opponent modeling within contextual reasoning.

In the \textbf{opponent model construction} stage, SOM builds an explicit opponent model grounded in Structural Causal Models (SCMs), which provide a structural framework to organize reasoning dependencies among observable factors and opponent's decisions. After each opponent action, the LLM performs reflection to infer how the observed outcome may have arisen---linking the opponent's decisions to contextual cues and hypothesizing intermediate reasoning variables that could explain this connection.  These insights are then used to progressively build and refine the SCMs, forming the explicit reasoning backbone.

In the \textbf{opponent prediction} stage, the LLM performs reasoning guided by the structured dependencies captured during the construction stage to anticipate the opponent's next action. 
At each step, the model draws on reasoning examples associated with the relevant dependency in the structure, which record prior successful inferences linking observed factors to opponent behavior. 
This allows the agent to continuously refine its reasoning with new observations, improving both the accuracy and adaptability of predictions in dynamic multi-agent interactions. 
 
Finally, we validate the effectiveness of our approach across multiple multi-agent game environments. Extensive experiments demonstrate that our framework significantly outperforms existing baseline methods when facing different opponents. Analysis of the training process further confirms that our method accurately learns opponent strategies during interactions.

Overall, our contributions to strategic reasoning can be summarized as follows:
\begin{itemize}
\item \normalsize  We propose \textbf{SOM}, a novel opponent modeling framework that leverages Structural Causal Models (SCMs) to transform opponent prediction into a structured and controllable reasoning process.

\item \normalsize  Within SOM, we implement two key mechanisms: a dynamic construction of the reasoning structure during interactions, and the integration of opponent-specific reasoning knowledge into the structured dependencies.

\item \normalsize  We empirically validate SOM across diverse multi-agent environments, showing that it outperforms strong baselines and adaptively captures the behavior of different opponents over time.
\end{itemize}

\section{Related work}
\subsection{Strategic Reasoning with LLMs }

Strategic reasoning~\cite{zhang2024llm} refers to the capability of an agent to analyze the opponent's history and the game state, infer the opponent's strategy and actions, and adjust its own strategy to select the best course of action based on these predictions. Early work like Cicero~\cite{meta2022human} combined language models with strategic reasoning, creating a conversational agent capable of playing Diplomacy. Cicero utilized an LLM to model other players' beliefs and intentions to predict their actions, enabling human-level play. Subsequent research has applied LLMs to various multi-player games.  In social deduction games like Werewolf~\cite{xu2023language, wu2024enhance}, studies aim to enhance agents' strategic abilities by enabling them to understand game mechanics and adapt to opponents' tactics, often involving implicit opponent prediction through dialogue analysis. Theory of Mind (ToM)~\cite{guo2023suspicion} and k-level thinking models~\cite{zhang2024k} have also been adapted to recursively infer opponents' hidden beliefs and predict their behavior in strategic reasoning. The EMO~\cite{yu2025llm}  method simulates opponent modeling by constructing multiple agent-specific models, but it still lacks an explicit representation of the opponent's decision-making process.

While these methods leverage the powerful reasoning capabilities of LLMs  and often incorporate some form of opponent action prediction, they typically treat opponent modeling as a general reasoning task. Although some approaches may use perspective-taking to simulate inferential processes, these often lack clear and controllable reasoning pathways.

\subsection{Structured Prompting for Reasoning }
Structured prompting, a technique that guides LLMs through multi-step reasoning by explicitly structuring the prompt format, has significantly enhanced their reasoning capabilities. A foundational approach is Chain-of-Thought (CoT)~\cite{wei2022chain}, which enables LLMs to generate a series of intermediate natural language reasoning steps. Building upon this, Self-Consistency (SC)~\cite{Wang2022SelfConsistencyIC} improves CoT's robustness by sampling diverse reasoning paths and aggregating results via majority voting. To overcome the inherent linearity of CoT, Tree-of-Thought (ToT)~\cite{yao2023tree} models reasoning as a tree-like exploration, allowing for branching and backtracking. Further generalizing this concept, Graph-of-Thought (GoT)~\cite{Besta2024GraphofThoughtsSE} employs arbitrary graph structures to represent complex dependencies between thoughts.
Building on this, Diagram-of-Thought (DoT)~\cite{Zhang2024OnTD} allows a single LLM to internally construct and reason over DAGs using role-specific tokens, streamlining multi-step reasoning without external control. Logic-of-Thought (LoT)~\cite{Li2025LogicofThoughtEL} further integrates formal logic into prompts to improve consistency and deductive precision.

While structured prompting has significantly enhanced the reasoning capabilities of LLMs, existing methods are predominantly designed for static problem settings and lack mechanisms to incorporate feedback or adapt their reasoning structures over time. As a result, they struggle to effectively capture opponent behavior in dynamic multi-agent environments characterized by strategic interactions and evolving behaviors. This limitation highlights the urgent need for approaches that enable more adaptive and opponent-aware reasoning in such settings.

\begin{figure*}[h]
    \centering

    \includegraphics[width=0.92\linewidth]{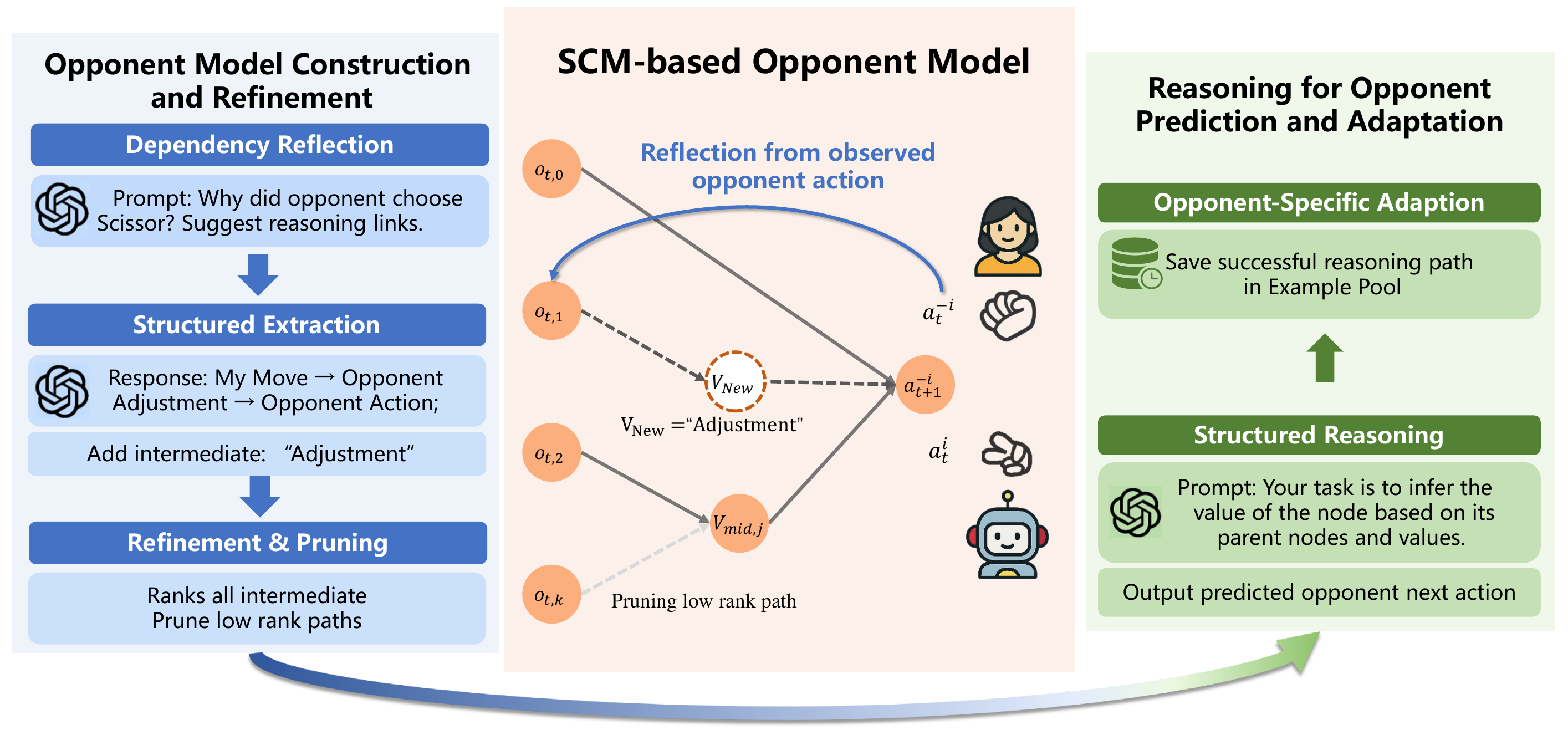}

    \caption{ Illustration of the opponent modeling pipeline of SOM. SOM operates in two explicit stages. First, it constructs the SCM representation of the opponent by building a structured causal graph that captures key decision-relevant variables and their dependencies. Second, it populates the structural relationships of this SCM using personalized reasoning examples derived from past interactions. During inference, SOM traverses the graph to simulate the opponent's reasoning process step by step, enabling explicit and adaptive opponent modeling. }
    \label{fig:main}
\end{figure*}

\subsection{Opponent Modeling}
Opponent modeling (OM), which analyzes and predicts other agents' behaviors in multi-agent systems, is a fundamental technique. To tackle unknown and non-stationary opponents: Encoder-decoder architectures~\cite{papoudakis2021agent} identify opponent models using only the controlled agent's local information. UAOM~\cite{yang2025uncertainty} captures aleatoric and epistemic uncertainties in stochastic opponent behaviors.Meta-learned Bayesian belief inference~\cite{zintgraf2021deep} combines variational autoencoders to model opponent beliefs; the meta-multiagent policy gradient theorem~\cite{kim2021policy} adapts to new agents by accounting for mutual non-stationary dynamics. GSCU~\cite{fu2022greedy} learns offline opponent policy embeddings and trains a universal best-response model. For diverse opponents, MBOM~\cite{yu2022model} simulates recursive reasoning via an environment model, adapting to various types by mixing improved policies. OEOM~\cite{jing2025open} continuously generates diverse opponents via population-based training and enhances robustness with in-context reinforcement learning. To exploit opponents: L2E~\cite{wu2022l2e} gains exploitation abilities through minimal interactions; M-FOS~\cite{lu2022model} achieves long-horizon shaping via model-free optimization; MOL~\cite{hu2023modeling} uses best response theory to approximate preferences for stable equilibrium improvements.
Unlike these traditional opponent modeling approaches, our work focuses on opponent modeling in LLM driven decision-making scenarios, which has not been adequately explored in existing research.

\section{Preliminaries}
\subsection{Partially Observable Stochastic Game}
\label{sec:posg}

We model the multi-agent interactions as a \textbf{Partially Observable Stochastic Game (POSG)}, a standard framework for sequential decision-making with multiple agents. A POSG is formally defined by the tuple~\cite{yang2020overview}:
\[
\langle N, S, \{A^i\}_{i=1}^N, P, \{R^i\}_{i=1}^N, \gamma, \{O^i\}_{i=1}^N, Q \rangle,
\]
where, $N$ is the set of agents, and $S$ denotes the state space. Each agent $i$ has an individual action space $A^i$, and the joint action space is defined as $A = \times_{i=1}^N A^i$. The state transition function is given by $P: S \times A \rightarrow \Delta(S)$, where $P(s' \mid s, a)$ denotes the probability of transitioning from state $s$ to state $s'$ after taking joint action $a$. Each agent $i$ receives a scalar reward determined by its reward function $R^i: S \times A \times S \rightarrow \mathbb{R}$, which gives a scalar reward for the transition $(s, a) \rightarrow s'$. $\gamma \in [0,1]$ is the discount factor. 

Each agent $i$ receives observations $o^i \in O^i$ from the environment, and the joint observation space is defined as $O = \times_{i=1}^N O^i$. The observation function $Q: S \times A \times S \rightarrow \Delta(O)$ specifies the probability of receiving a joint observation $o$ given joint action $a$ and next state $s'$, i.e., $Q(o \mid a, s')$.

The agent's local history at time $t$ is the sequence of its past observations, actions, and rewards: $h_t^i = (o_0^i, a_0^i, r_0^i, \dots, a_{t-1}^i, r_{t-1}^i, o_t^i)$.\\ The agent's policy maps this history to a distribution over actions: \\$\pi^i(a_t^i \mid h_t^i)$.

In this work, we focus on the perspective of the self-agent (the agent under our control), denoted by superscript $i$. All other agents, collectively denoted by $-i$, are treated as opponents. Each opponent's policy $\pi^{-i}$ is sampled from a predefined and diverse policy set $\Pi^{\text{opp}}$, which includes fixed, rule-based, and adaptive strategies.

During adaptation, the self-agent $i$ interacts repeatedly with opponents over $M$ episodes of POSG. The objective is to derive a policy $\pi^i$ that maximizes the expected cumulative reward over the time horizon $T$ and across all $M$ episodes:
\[
\max_{\pi^i} \mathbb{E}_{\pi^{-i} \sim \Pi^{\text{opp}}} \left[\sum_{m=1}^{M} \sum_{t=0}^{T} R_{t}^i \right].
\]

\subsection{Structural Causal Models}

A \textbf{Structural Causal Model (SCM)} \cite{pearl2000causality} provides a formal framework for representing causal relationships, comprising a set of variables, a causal graph, and structural equations.

\paragraph{Causal Graph.}
The causal relationships among variables are represented by a Causal Graph, denoted as $\mathcal{G}(\mathcal{V}, \mathcal{E})$.
\begin{itemize}
    \item $\mathcal{V}$ is a set of variables (nodes) in the model.
    \item $\mathcal{E}$ is a set of directed edges, where an edge $V_i \to V_j$ signifies that $V_i$ is a direct causal factor for $V_j$.
\end{itemize}
The graph $\mathcal{G}$ is a Directed Acyclic Graph (DAG), ensuring no causal cycles.

\paragraph{Structural Equations.}
For each variable $V_j \in \mathcal{V}$, its value is determined by its direct causal parents---denoted as $Pa(V_j)$---which are the set of variables in $\mathcal{V}$ with directed edges pointing to $V_j$, along with an exogenous disturbance variable $U_j$ that accounts for external influences not explained by the model. Each such relationship is captured by a structural function $f_j$:
\[
V_j = f_j(Pa(V_j),\ldots,U_j).
\] 
These structural functions $f_j$ define the mechanism by which the value of each variable is determined by its direct causes.

In our work, we adopt SCM to formalize the opponent's decision-making process. The variables $V$ encompass not only observable states, but also crucial latent variables representing the opponent's internal state (e.g., beliefs). The causal graph $\mathcal{G}$ structures the reasoning flow from observations to beliefs and then to actions, with each step governed by a structural function $f_j$ that represents a decision process. While the  graph and functions are unknown, the core premise of our work is that they can be dynamically inferred and approximated by LLMs. Our framework, SOM, is designed to instantiate this SCM, using an LLM to both discover the causal structure and execute the reasoning within it.

\section{Method}

To overcome the limitations of unstructured opponent modeling, we propose SOM, a framework that grounds the modeling process in Structural Causal Models and enables structured reasoning from observations to opponent actions.
\vspace{-0.0em}

\subsection{Overview of SOM Framework}

In multi-agent environments, a self-agent's success critically hinges on accurately predicting an opponent's next action based on its own observations and interaction history. However, achieving precision and adaptively accurate opponent modeling in dynamic settings remains a significant challenge.

To address this, we propose SOM, a novel framework that leverages the principles of SCMs for precision and adaptive opponent behavior prediction. SOM's overall architecture is designed to enhance modeling adaptability by uncovering the underlying logic of an opponent's decisions.

As shown in Figure~\ref{fig:main}, the framework consists of two interconnected mechanisms: Dynamic SCM Construction and Refinement, which builds and updates the structured representation of the opponent's decision process through a causal graph; and Reasoning for Opponent Prediction and Adaptation, which performs structured inference within this SCM using personalized reasoning knowledge to predict opponent actions. The following sections detail these two components and their interactions.

\subsection{Dynamic SCM Construction and Refinement}

To systematically establish a structured link between observations and opponent behavior, SOM grounds opponent model construction in the framework of Structural Causal Models (SCMs), where this structured dependency is explicitly represented through a causal graph $\mathcal{G}(\mathcal{V}, \mathcal{E})$. Accordingly, SOM dynamically constructs and continuously refines this graph to capture how observable factors and latent reasoning variables jointly shape the opponent's decisions over time. The process follows an "observation---reflection---extraction---consolidation---pruning" cycle, enabling the model to adaptively update its representation of the opponent's decision logic as interactions unfold.  

\textbf{Graph Initialization.}
SOM begins by constructing a minimal directed acyclic graph $\mathcal{G}_0 = (\mathcal{V}_0, \mathcal{E}_0)$ before interaction. The initial node set $\mathcal{V}_0$ includes all observable variables $\{o^i_{t,1}, \dots, o^i_{t,k}\}$ and the opponent's action $a^{-i}_t$. Edges are added from each observation variable to the opponent action $\mathcal{E}_0 = \{ (o^i_{t,k}, a^{-i}_t) \mid  \forall k \}$, representing an initial hypothesis that all observations may directly influence the opponent's action.

\textbf{Reflection Phase.}
As the interaction proceeds, after observing the opponent's actual action $a^{-i}_t$ in each round, SOM prompts the LLM to generate a natural language reflection. This reflection, based on the interaction history, the agent's current observation $o^i_t$, and the opponent's action $a^{-i}_t$, hypothesizes potential intermediate reasoning steps or latent beliefs that led the opponent from observations to the final action.

\textbf{Structured Extraction.}
Another LLM module parses this reflective text, transforming unstructured natural language into structured causal chains. This process extracts intermediate nodes $V_{\text{mid}}$ that lie between observations and actions, along with the specific causal pathways they form (e.g., $o^i_{t,k} \to V_{\text{mid}} \to a^{-i}_t$).

\textbf{Graph Update and Consolidation.}
After extraction, the system executes Graph Update and Consolidation. For each newly extracted intermediate node $V_{\text{new}}$, SOM queries an LLM to determine if it is semantically equivalent to any existing node in the graph's node set $\mathcal{V}$. To do this, the LLM receives the description of $V_{\text{new}}$ and a list of all existing nodes in $\mathcal{V}$, and then makes a matching decision. Concurrently, the system maintains a reinforcement count $c(V)$ for each intermediate node $V \in \mathcal{V}$: if $V_{\text{new}}$ matches an existing node $V_{\text{exist}}$, its count is incremented ($c(V_{\text{exist}}) \leftarrow c(V_{\text{exist}}) + 1$). If no match is found, $V_{\text{new}}$ is added to $\mathcal{V}$ as a new node with $c(V_{\text{new}}) = 1$, and the edge set $\mathcal{E}$ is updated according to the extracted causal chain.

\textbf{Graph Refinement and Pruning.}
To control complexity and retain critical causal hypotheses, the framework performs Graph Expansion and Refinement. After each update, SOM ranks all intermediate nodes based on their reinforcement counts $c(V)$ and retains only the top-$K$ nodes. Nodes below this rank are pruned from the graph, ensuring the graph remains concise and effective by preserving repeatedly validated decision logic and discarding transient or refuted hypotheses.
\vspace{-1.0em}

\subsection{Reasoning for Opponent Prediction and Adaptation}

Given the constructed SCM that represents the opponent's decision process, SOM performs opponent prediction through structured reasoning along the dependencies encoded in the model. Specifically, reasoning proceeds in three stages—topological inference, example-guided reasoning, and personalized adaptation. By simulating the functional relationships among variables defined in the SCM, SOM predicts the opponent's next action $a^{-i}_{t+1}$ and continually updates its understanding of the opponent as interactions unfold.

SOM traverses the causal graph $\mathcal{G}$ in topological order, ensuring that each node $V_j$ is inferred only after all of its parent nodes $Pa(V_j)$ have been determined. The root nodes, typically the agent's observations $o^i_{t+1}$, obtain their values directly from the environment, whereas intermediate and action nodes are computed through a step-by-step inference process that depends on their parent nodes. For each node $V_j$, its value is determined by a structural equation $V_j = f_j(Pa(V_j))$, which is implemented by an LLM equipped with dynamically updated knowledge to simulate the parent-to-child causal mapping.

To enable accurate node inference, SOM constructs a tailored prompt for the LLM. The prompt consists of two components: (i) the current inferential context, namely the determined values of all parent nodes $Pa(V_j)$, and (ii) relevant reasoning examples retrieved from an opponent-specific example pool\textbf{ $\mathcal{P}_{\text{opponent}}$}. The retrieval process first converts the parent nodes and their values into a textual query, and then performs semantic-similarity search in $\mathcal{P}$ to identify the top-$M$ most similar examples. Combining the context with these retrieved examples, the LLM performs example-guided reasoning to infer the most likely value of $V_j$ and generate the corresponding reasoning text. This personalized process enhances prediction stability.

\begin{algorithm}[t]
\caption{SOM Opponent Modeling Loop}
\label{alg:SOM}
\begin{algorithmic}[1]
\State \textbf{Initialize:} Minimal Causal Graph $\mathcal{G}$; Example Pool $\mathcal{P}_{\text{opponent}} \leftarrow \emptyset$
\For{each interaction round $t=1$ to $T$}
\State Observe current observation $o^i_t$ and opponent's actual action $a^{-i}_{t}$
    \State \textbf{Construct/Update Graph:} Prompt LLM to hypothesize causal links from $o^i_t$ to $a^{-i}_{t}$
    \State Extract nodes and edges to update the graph $\mathcal{G}$
    \State \textbf{Update Example Pool:} If prediction for round $t-1$ ($\hat{a}^{-i}_{t-1}$) was correct, add its successful parent-to-child reasoning steps to $\mathcal{P}_{\text{opponent}}$
    \State \textbf{Predict Next Action:} Traverse $\mathcal{G}$ in topological order
    \For{each node $V_j$ in topological order}
        \State Retrieve examples based on parent nodes $Pa(V_j)$
        \State Infer node value $V_j = f_j(Pa(V_j), \text{examples})$
    \EndFor
    \State Output predicted action $\hat{a}^{-i}_{t+1} = \text{value of } V_{a^{-i}}$
    \State Agent selects own action based on $\hat{a}^{-i}_{t+1}$
\EndFor

\end{algorithmic}
\end{algorithm}

SOM maintains a shared causal graph $\mathcal{G}$ while achieving personalized adaptation to different opponents through dynamically maintained, opponent-specific example pools. For each opponent, a distinct example pool $\mathcal{P}_{\text{opponent}}$ is incrementally populated with parent-to-child reasoning steps generated by the LLM (only when the predictions are correct). Each example $e$ is formally represented as a four-tuple, $e = \langle \text{parent values}, \text{child value}, \text{reasoning text},\\ \text{target link} \rangle$, capturing one validated reasoning event. A strict credit-assignment policy ensures the quality of the stored knowledge: only when the predicted action $\hat{a}^{-i}_{t+1}$ matches the observed action $a^{-i}_{t+1}$ are all intermediate reasoning steps accepted, formatted as examples, and stored in the corresponding pool. This mechanism accumulates high-quality, opponent-specific reasoning knowledge, enabling SOM to perform highly personalized and adaptive opponent modeling. The complete process of SOM is detailed in Algorithm 1.

\section{Experiment}

\begin{table*}[h]
\centering
\caption{Win rates of different reasoning methods against various opponents in the G0.8A game. 
Rows represent the evaluated agent, and columns represent the opponent type. 
SOM achieves the highest overall average win rate, particularly excelling against the Mixed opponent group that aggregates diverse reasoning strategies.}

\begin{tabular}{l|ccccccc|c}
\hline
\multirow{2}{*}{\textbf{Evaluated Method}} 
& \multicolumn{7}{c|}{\textbf{Opponent Method}} 
& \multirow{2}{*}{\textbf{Avg.}} \\ 
\cline{2-8}
 & \textbf{LLM only} & \textbf{CoT} & \textbf{ToT} & \textbf{K-R} & \textbf{Reflexion} & \textbf{Ours} &  \textbf{Mixed} & \\ 
\hline
LLM only  & 0.19 & 0.04 & 0.12 & 0.02 & 0.10 & 0.03 & 0.07 & 0.08 \\
CoT~\cite{wei2022chain}       & 0.68 & 0.16 & 0.54 & 0.28 & 0.32 & 0.09 & 0.36 & 0.35 \\
ToT~\cite{yao2023tree}        & 0.46 & 0.18 & 0.22 & 0.12 & 0.22 & 0.11 & 0.26 & 0.22 \\
K-R~\cite{zhang2024k}         & \textbf{0.84} & 0.54 & 0.48 & 0.24 & 0.45 & 0.17 & 0.42 & 0.45 \\
Reflexion~\cite{shinn2023reflexion} & 0.64 & 0.10 & 0.40 & 0.20 & 0.26 & \textbf{0.23} & 0.54 & 0.34 \\
Ours      & 0.80$\pm$\scriptsize{0.14} & \textbf{0.61}$\pm$\scriptsize{0.11} & \textbf{0.59}$\pm$\scriptsize{0.09} & \textbf{0.39}$\pm$\scriptsize{0.12} & \textbf{0.47}$\pm$\scriptsize{0.08} & 0.19$\pm$\scriptsize{0.10} & \textbf{0.64}$\pm$\scriptsize{0.13} & \textbf{0.53} \\ \hline

\end{tabular}
\label{tab:g08a}
\end{table*}

\subsection{ Experiment Setup}
\textbf{Environments.} We evaluate our approach in three distinct multi-agent game environments:
\begin{itemize}
    \item \textbf{G08A} \cite{zhang2024k}: A multi-round number-guessing game in which players choose a number between 1 and 100 in each round, aiming to be closest to 80\% of the group average. This is a variant of the classic "Guess 2/3 of the Average" game proposed by Ledoux \cite{ledoux1981concours}, where success hinges on accurately anticipating others' choices.
    \item \textbf{Survival Auction Game (SAG)} \cite{mao2024alympics}: A multi-round sealed-bid auction game, adapted from the classic sealed-bid auction game \cite{vickrey1961counterspeculation}, where players bid for water to restore health points. In each round, players submit bids privately, and the highest bidder wins the water. Success hinges on accurately anticipating opponents' bids to acquire water at the lowest possible cost.
     \item \textbf{Undercover Game} \cite{xu2023magic}:  A social deduction game where players are Civilians or Undercovers with different words. Players infer their own roles from clues. Civilians aim to identify Undercovers, who try to conceal their roles. The core is reasoning about others' roles based on their behaviors. 

\end{itemize}
\textbf{Enhanced Reasoning Baselines.}
Recent advances in prompting techniques have significantly improved the reasoning capabilities of large language models. We focus on four representative baseline methods:
\begin{itemize}
    \item \textbf{Chain of Thought (CoT)} \cite{wei2022chain} is a prompting method that guides LLMs to generate explicit intermediate reasoning steps, enabling them to decompose complex problems into simpler parts.
    \item \textbf{Tree of Thoughts (ToT)} \cite{yao2023tree} generalizes CoT by allowing LLMs to explore multiple reasoning paths. It facilitates deliberate decision-making through evaluating multiple reasoning paths, self-evaluating progress, and applying lookahead and backtracking strategies.
    \item \textbf{K-Level Reasoning (K-R)} \cite{zhang2024k} equips LLMs with recursive strategic reasoning, enabling agents to form higher-order beliefs about others' beliefs and adapt dynamically in multi-agent environments.
    \item \textbf{Reflexion} \cite{shinn2023reflexion} enables LLM agents to improve through linguistic feedback instead of weight updates, by verbally reflecting on task feedback and storing reflections in episodic memory for better future decisions. 
\end{itemize}

Meanwhile, we introduce an additional baseline named Mixed Opponent (\textbf{Mixed}), which is composed by randomly sampling opponent behaviors from CoT, ToT, K-R, and Reflexion agents. This baseline is designed to simulate a more diverse and uncertain opponent environment.

\begin{table*}[t]
\centering
\caption{Average survival rounds of different reasoning methods against various opponents in the Survival Auction Game (SAG). 
Rows denote the evaluated reasoning method, while columns denote the opponent type. 
SOM achieves the longest survival across most opponent types, especially under the Mixed setting, demonstrating its robustness and adaptability in dynamic auction interactions.}
\begin{tabular}{l|ccccccc|c}
\hline
\multirow{2}{*}{\textbf{Evaluated Method}} 
& \multicolumn{7}{c|}{\textbf{Opponent Method}} 
& \multirow{2}{*}{\textbf{Avg.}} \\ 
\cline{2-8}
 & \textbf{LLM only} & \textbf{CoT} & \textbf{ToT} & \textbf{K-R} & \textbf{Reflexion} & \textbf{Ours} &  \textbf{Mixed} & \\ 
\hline
\textbf{LLM only}  & 5.7 & 4.0 & 5.4 & 4.7 & 6.8 & 4.2 & 4.9 & 5.1 \\
\textbf{CoT}~\cite{wei2022chain}       & 6.5 & 5.0 & 7.8 & 5.6 & 5.6 & 4.9 & 5.5 & 5.8 \\
\textbf{ToT}~\cite{yao2023tree}       & 6.0 & 5.8 & 3.7 & 5.1 & 6.4 & 5.5 & 4.6 & 5.3 \\
\textbf{K-R}~\cite{zhang2024k}       & 8.1 & 8.4 & 7.8 & 5.6 & 7.4 & \textbf{6.1} & 6.2 & 7.1 \\
\textbf{Reflexion}~\cite{shinn2023reflexion} & 3.7 & 3.7 & 7.2 & 6.6 & 4.4 & 5.8 & 5.2 & 5.2 \\
\textbf{Ours}      & \textbf{9.1}$\pm$\scriptsize{0.73} & \textbf{8.8}$\pm$\scriptsize{0.80} & \textbf{8.3}$\pm$\scriptsize{0.69} & \textbf{7.9}$\pm$\scriptsize{0.81} & \textbf{8.1}$\pm$\scriptsize{0.80} & 4.9$\pm$\scriptsize{0.72} & \textbf{7.4}$\pm$\scriptsize{0.83} & \textbf{7.8} \\ \hline

\end{tabular}
\label{tab:sag}

\end{table*}
For a fair comparison, all methods are provided with a warm-up phase of 5 episodes prior to evaluation, during which interaction histories are collected. These histories are supplied as contextual input to the baseline methods during  evaluation. During the evaluation phase, no additional cross-episode history is provided.
Similarly, our method also fixes the SCM structure during evaluation and does not perform any cross-episode updates or adaptation, in order to ensure consistency and reproducibility across multi-runs.
Unless otherwise specified, all methods and results are evaluated using GPT-4o as the base model.

More detailed experimental settings can be found in the supplementary materials.

\subsection{Results}

\paragraph{\textbf{G0.8A Game and Survival Auction Game}}Tables~\ref{tab:g08a} and \ref{tab:sag} summarize performance across G0.8A and SAG environments. In G0.8A, SOM achieves the highest average win rate (0.53). Against Mixed opponents, SOM substantially outperforms ToT and CoT, demonstrating its adaptability to heterogeneous strategies. While K-R excels against single LLM-only opponents, its performance declines against diverse groups, showing the limitations of fixed $k$-level assumptions under non-stationary behaviors. Reflexion shows moderate gains but slightly higher win rates when SOM is the opponent; this asymmetry reflects that SOM’s stable, equilibrium-oriented reasoning may increase tie frequency in a game where win rates are theoretically low (0.2–0.3). In contrast, ToT and CoT struggle with dynamic mixed strategies, confirming that explicit two-stage modeling provides a tangible advantage.

In SAG, SOM consistently leads all baselines with an average survival of 7.8 rounds. Notably, SOM surpasses K-R in nearly all matchups, highlighting its advantage when optimal bidding requires continuous adjustment. CoT and ToT exhibit variable performance, with ToT struggling against heterogeneous strategies, underscoring the limitations of static tree-based reasoning. While Reflexion improves via episodic feedback, it lacks SOM's stability due to the absence of structured model construction. Across both environments, SOM’s two-stage approach—separating model construction from prediction and integrating opponent-specific knowledge—demonstrates robust adaptability and superior performance against diverse or dynamically changing opponents.

\begin{figure*}[t]
    \centering
     \begin{subfigure}[b]{0.49\linewidth}
        \centering
        \includegraphics[width=0.95\linewidth]{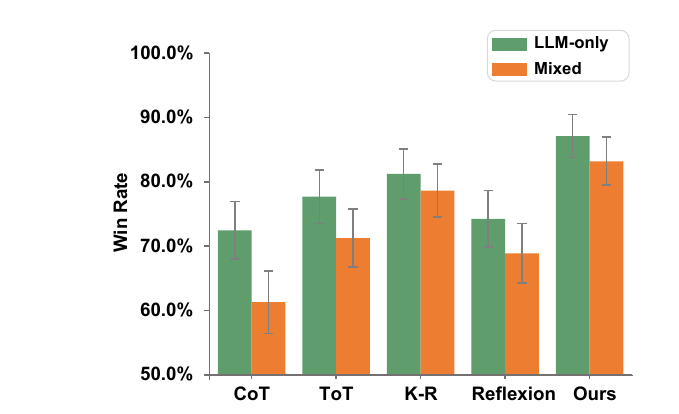}
        \captionsetup{margin={0pt, 100pt}, singlelinecheck=off}
        \caption{As Civilian in the game}
        \label{fig:witu1}
    \end{subfigure}
    \begin{subfigure}[b]{0.49\linewidth}
        \centering
        \includegraphics[width=1.00\linewidth]{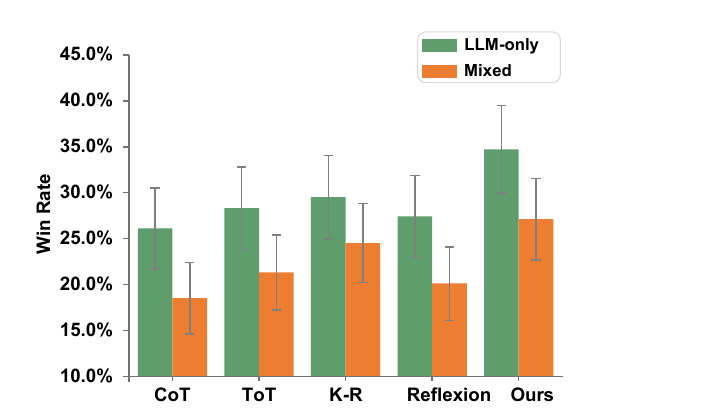}
        \caption{As Undercover in the game}
        \label{fig:witu2}
    \end{subfigure}
        \vspace{-1em}
    \caption{Win Rate against different opponents in Undercover game. The performance of SOM and baseline methods is evaluated against LLM-only opponents and Mixed opponents. SOM consistently outperforms all baselines in both scenarios.
    }
    \label{fig:witu}
        \vspace{-1.0em}
\end{figure*}

\paragraph{\textbf{Undercover Game.}} 
In the Undercover Game (Figure~\ref{fig:witu}), which requires linguistic reasoning and implicit role inference, SOM again demonstrates consistent superiority over all baselines. It achieves higher win rates both as a Civilian (Figure~\ref{fig:witu1}), when identifying deceptive language patterns, and as an Undercover (Figure~\ref{fig:witu2}), when strategically concealing its role.
This performance improvement highlights SOM's ability to integrate structural knowledge about discourse patterns—such as topic shifts and semantic divergence—into its reasoning process. While CoT and ToT often overfit to surface-level linguistic cues, SOM's structured model allows it to capture how utterances functionally depend on hidden role intent.

\paragraph{\textbf{Multi-Round Interaction Analysis}} To investigate SOM's long-term performance, we analyze its prediction deviation and win rate in G0.8A over continuous episodes. Both SOM and Reflexion are initialized with full historical context—SOM through state refinement and Reflexion via retrieved historical reflections—against LLM-only opponents.

\begin{figure}[t]
    \centering

    \begin{subfigure}[b]{0.49\linewidth}
        \centering
        \includegraphics[width=1.00\linewidth]{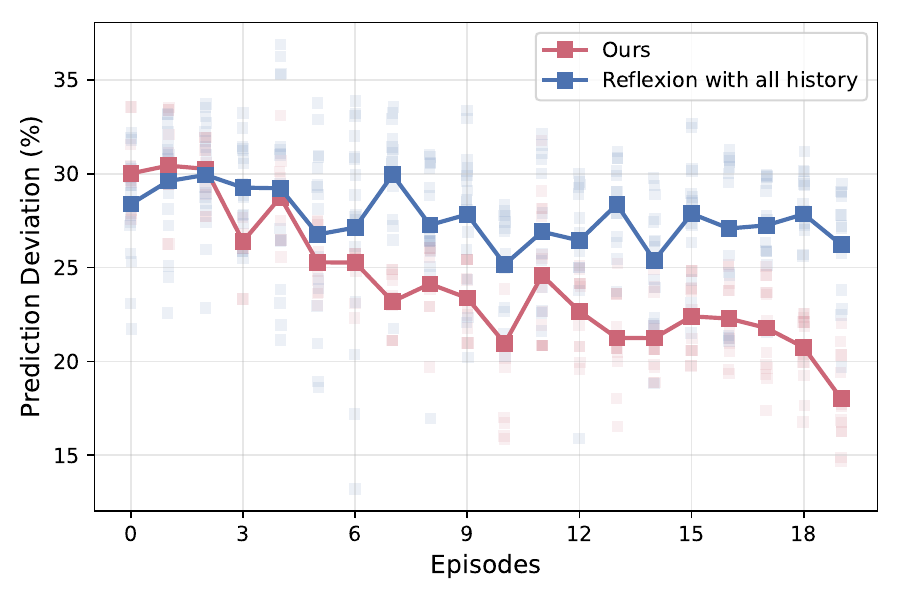}
        \caption{Prediction deviation.}
        \label{fig:sub1}
    \end{subfigure}
    \begin{subfigure}[b]{0.49\linewidth}
        \centering
        \includegraphics[width=1.00\linewidth]{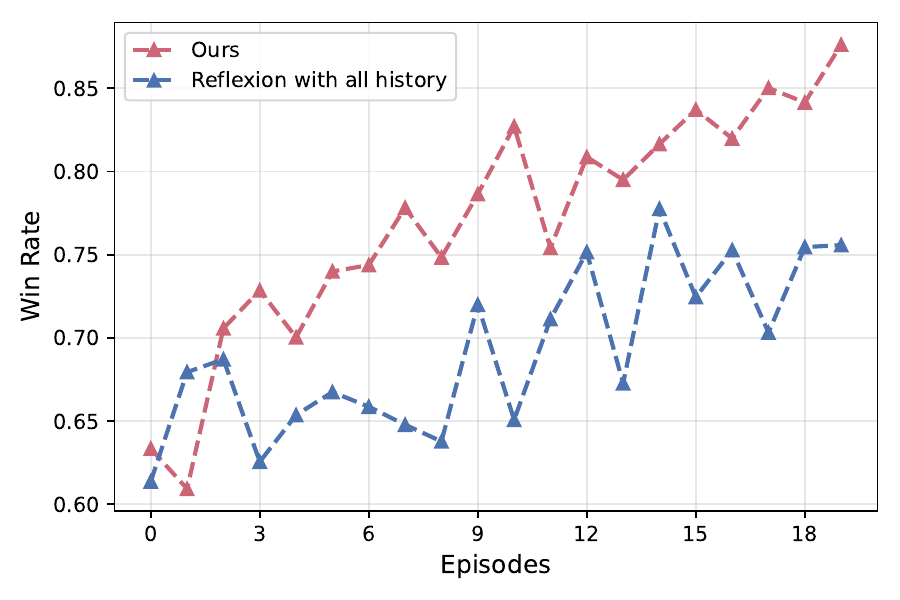}
        \caption{Win rate over episodes.}
        \label{fig:sub2}
    \end{subfigure}
        \vspace{-1em}

    \caption{Action prediction deviation and win rate over episodes in G0.8A.
(a) Prediction Deviation: SOM maintains higher accuracy and stability than Reflexion. (b) Win Rate: SOM exhibits superior learning progress and a higher final win rate over extended episodes.}
    \label{fig:train1}
    \vspace{-1em}
\end{figure}

As shown in Figure~\ref{fig:sub1}, SOM's prediction deviation steadily decreases and stabilizes below Reflexion, indicating that the dynamic SCM refinement effectively captures opponent patterns. This superior accuracy translates into a strategic advantage: Figure~\ref{fig:sub2} shows SOM's win rate consistently increases in tandem with error reduction, eventually significantly outperforming all baselines. Conversely, Reflexion plateaus due to the absence of a structured, continuous modeling process. These results validate SOM’s core design—improving decision quality through adaptive opponent modeling—and demonstrate its robust capacity for progressive reasoning and adaptation.

\setlength{\tabcolsep}{1pt}
\begin{table}[t!]
\centering
\caption{Ablation study of SOM components. We incrementally add SOM's core modules to evaluate their impact on Prediction Deviation  and Win Rate  in the G0.8A game. 
}
\label{tab:ablation}
\begin{tabularx}{\columnwidth}{>{\raggedright\arraybackslash}X c c c}
\toprule
\textbf{Model Variant} 
  & \textbf{\makecell{Prediction\\Deviation}} 
  & \textbf{$\downarrow$ (\%)\quad} 
  & \textbf{Win Rate $\uparrow$} \\
\midrule
LLM-only                 & 43.0  & \quad & 0.04 \\
+ Static Graph           & 30.4  & \quad & 0.19 \\
+ Intermediate Nodes     & 27.1  & \quad & 0.51 \\
+ Graph Refine           & 26.9  & \quad & 0.54 \\
\textbf{\makecell[l]{+ Reasoning Examples \\ (SOM)}} & \textbf{25.3} & \quad & \textbf{0.61} \\
\bottomrule
\end{tabularx}
\vspace{-1.0em}
\end{table}

\subsection{Analysis of SOM's Components }

\subsubsection{\textbf{Ablation Study}} 
To validate the effectiveness of each core component of SOM, we conduct a series of ablation experiments by incrementally adding key modules and evaluating their impact on model performance. The experiments are carried out in the G0.8A game environment, and the results are summarized in Table~\ref{tab:ablation}.

\textbf{LLM-only:} As the baseline setting, this variant involves no structured modeling. It yields the highest prediction deviation (43.0\%) and the lowest win rate, indicating the limitations of relying solely on end-to-end language model reasoning without structural guidance.

\textbf{+ Static Graph:} When a static causal graph is introduced, consisting only of direct edges from observation nodes to the action node—the prediction deviation drops significantly, and win rate improves. This demonstrates that even a basic hypothesized reasoning structure can provide meaningful guidance for the LLM's inference process, improving both stability and directionality.

\textbf{+ Intermediate Nodes:} We then incorporate the mechanism for dynamically extracting intermediate variables from LLM-generated reflections. This leads to a substantial boost in both prediction accuracy and win rate, highlighting that the key to effective modeling lies not merely in surface-level observation-action mappings, but in uncovering intermediate reasoning steps that reflect the opponent's underlying decision process. These steps guide the LLM to reason in a structured, step-by-step manner.

\textbf{+ Graph Refine:} The addition of the graph refinement and pruning mechanism---based on reinforcement counts---helps retain reliable causal paths and eliminate spurious or hallucinated connections. This further stabilizes performance by reducing redundancy and preserving the most consistent decision logic.

\textbf{+ Reasoning Examples (SOM):} Finally, we enable the full SOM framework by adding the personalized example-pool mechanism. This module retrieves and leverages previously successful reasoning trajectories that are semantically similar to the current context, effectively simulating the structural equations $V_j = f_j(Pa(V_j))$ defined in the SCM framework. This step validates the central advantage of our two-stage design: while the causal graph captures the general structure of an opponent's decision logic, the reasoning examples instantiate the functional mappings within that structure, enabling highly personalized and adaptive opponent modeling. This results in the lowest prediction deviation (25.3\%) and the highest win rate (0.61) among all variants.

\setlength{\tabcolsep}{1pt}
\begin{table}[t!]
\centering
\caption{Knowledge transfer test. Each agent plays against the same strong opponent: CoT (GPT-4o). \textbf{SOM-T} denotes a transferred SOM model originally constructed by a GPT-4o agent during its interaction with the CoT (GPT-4o) opponent.}
\label{tab:transfer}
\begin{tabularx}{\columnwidth}{l c l c}
\toprule
\textbf{Agent Variant} 
  & \textbf{\makecell{Prediction\\Deviation}} 
  & \textbf{$\downarrow$ (\%)\quad } 
  & \textbf{Win Rate $\uparrow$} \\
\midrule
GPT-4o + SOM       & 25.3          &\quad & 0.61 \\
\midrule
LLaMA-3-8B + CoT    & 76.1          &\quad & 0.07 \\
LLaMA-3-8B + SOM-T & \textbf{45.8} &\quad & \textbf{0.31} \\
\midrule
Mixtral-8B + CoT    & 95.6          &\quad & 0.02 \\
Mixtral-8B + SOM-T & \textbf{65.2} &\quad & \textbf{0.27} \\
\bottomrule
\end{tabularx}
\vspace{-1.0em}
\end{table}
Overall, the ablation results clearly demonstrate that each component of SOM contributes significantly to its final performance. In particular, the introduction of intermediate variables and the use of personalized reasoning examples are critical to improving both predictive accuracy and strategic decision-making.

\subsubsection{\textbf{Structured Knowledge Transfer Analysis}}
One core advantage of SOM is its ability to construct structured opponent models that generalize across different LLMs. To validate this, we conduct a knowledge transfer experiment (Table~\ref{tab:transfer}), testing whether the SOM-generated model—comprising a causal graph $\mathcal{G}$ and an opponent-specific example pool $\mathcal{P}_{\text{opponent}}$—can be effectively reused by other agents.

We begin by allowing a strong agent (GPT-4o + SOM) to interact with a strong opponent (CoT-driven GPT-4o) and store the final constructed opponent model, including the causal graph and the reasoning examples. Then, we test two weaker open-source models (LLaMA-3-8B and Mixtral-8B) by directly loading this constructed model (denoted as \textbf{SOM-T}) and using it to play against the same CoT (GPT-4o) opponent.

As shown in Table~\ref{tab:transfer}, SOM-T substantially improves the performance of weaker models without any additional training. When LLaMA-3-8B uses the transferred SOM model, its prediction deviation falls from 76.1\% to 45.8\% and its win rate rises from 0.07 to 0.31. Mixtral-8B shows the same pattern.
Importantly, the improvement is achieved by directly loading the SOM-based opponent model into the target agent, without fine-tuning the target LLM. This demonstrates that the structured representation learned by SOM encodes reusable behavioral regularities that benefit different model architectures. At the same time, transferred models do not fully close the gap to the high-capacity SOM instantiation, indicating that recipient model capacity and inference ability still constrain final performance. In short, SOM produces structured opponent knowledge that is transferable and practically useful across LLMs, while remaining complementary to improvements in base model capacity.
\section{Conclusion}

We introduce \textbf{SOM}, a novel two-stage opponent modeling framework inspired by structural causal modeling principles. By dynamically constructing and refining a structured reasoning graph, SOM explicitly decouples the process of opponent model construction from that of behavior prediction. Comprehensive experiments demonstrate that this structured reasoning approach substantially improves prediction accuracy and decision-making performance across diverse game-theoretic environments. Furthermore, the modular opponent models produced by SOM can be seamlessly transferred to empower other agents, highlighting its generality and practical utility.

Nevertheless, we acknowledge a limitation: the "causal" structures discovered by SOM represent  functional dependencies inferred from observational data, rather than verified causal mechanisms of the opponent's cognition. Bridging this gap requires integrating more principled causal discovery techniques or controlled interaction settings in future work.

Overall, SOM offers a promising step toward building LLM-based agents that are more adaptive, interpretable, and robust in complex multi-agent environments.

\begin{acks}
This work was supported by the National Science and Technology Major Project under Grant No. 2022ZD0116403, and in part by the Beijing Natural Science Foundation under Grant No. 4264131.

\end{acks}
\balance
\bibliographystyle{ACM-Reference-Format} 
\bibliography{sample}

@article{bianchi2024well,
  title={How well can llms negotiate? negotiationarena platform and analysis},
  author={Bianchi, Federico and Chia, Patrick John and Yuksekgonul, Mert and Tagliabue, Jacopo and Jurafsky, Dan and Zou, James},
  journal={arXiv preprint arXiv:2402.05863},
  year={2024}
}

@techreport{horton2023large,
  title={Large language models as simulated economic agents: What can we learn from homo silicus?},
  author={Horton, John J},
  year={2023},
  institution={National Bureau of Economic Research}
}

@inproceedings{li2024econagent,
  title={Econagent: large language model-empowered agents for simulating macroeconomic activities},
  author={Li, Nian and Gao, Chen and Li, Mingyu and Li, Yong and Liao, Qingmin},
  booktitle={Proceedings of the 62nd Annual Meeting of the Association for Computational Linguistics (Volume 1: Long Papers)},
  pages={15523--15536},
  year={2024}
}

@article{chen2024comm,
  title={CoMM: Collaborative multi-agent, multi-reasoning-path prompting for complex problem solving},
  author={Chen, Pei and Han, Boran and Zhang, Shuai},
  journal={arXiv preprint arXiv:2404.17729},
  year={2024}
}

@article{carroll2019utility,
  title={On the utility of learning about humans for human-ai coordination},
  author={Carroll, Micah and Shah, Rohin and Ho, Mark K and Griffiths, Tom and Seshia, Sanjit and Abbeel, Pieter and Dragan, Anca},
  journal={Advances in neural information processing systems},
  volume={32},
  year={2019}
}

@inproceedings{yu2025llm,
  title={LLM-Based Explicit Models of Opponents for Multi-Agent Games},
  author={Yu, XiaoPeng and Zhang, Wanpeng and Lu, Zongqing},
  booktitle={Proceedings of the 2025 Conference of the Nations of the Americas Chapter of the Association for Computational Linguistics: Human Language Technologies (Volume 1: Long Papers)},
  pages={892--911},
  year={2025}
}

@article{sun2023head,
  title={Head-to-tail: How knowledgeable are large language models (llm)? AKA will llms replace knowledge graphs?},
  author={Sun, Kai and Xu, Yifan Ethan and Zha, Hanwen and Liu, Yue and Dong, Xin Luna},
  journal={arXiv preprint arXiv:2308.10168},
  year={2023}
}

@article{rasal2024llm,
  title={Llm harmony: Multi-agent communication for problem solving},
  author={Rasal, Sumedh},
  journal={arXiv preprint arXiv:2401.01312},
  year={2024}
}

@article{zhang2024llm,
  title={LLM as a Mastermind: A Survey of Strategic Reasoning with Large Language Models},
  author={Zhang, Yadong and Mao, Shaoguang and Ge, Tao and Wang, Xun and de Wynter, Adrian and Xia, Yan and Wu, Wenshan and Song, Ting and Lan, Man and Wei, Furu},
  journal={arXiv preprint arXiv:2404.01230},
  year={2024}
}

@article{imani2023mathprompter,
  title={Mathprompter: Mathematical reasoning using large language models},
  author={Imani, Shima and Du, Liang and Shrivastava, Harsh},
  journal={arXiv preprint arXiv:2303.05398},
  year={2023}
}

@article{nashed2022survey,
  title={A survey of opponent modeling in adversarial domains},
  author={Nashed, Samer and Zilberstein, Shlomo},
  journal={Journal of Artificial Intelligence Research},
  volume={73},
  pages={277--327},
  year={2022}
}

@article{meta2022human,
  title={Human-level play in the game of Diplomacy by combining language models with strategic reasoning},
  author={{Meta Fundamental AI Research Diplomacy Team (FAIR)} and Bakhtin, Anton and Brown, Noam and Dinan, Emily and Farina, Gabriele and Flaherty, Colin and Fried, Daniel and Goff, Andrew and Gray, Jonathan and Hu, Hengyuan and others},
  journal={Science},
  volume={378},
  number={6624},
  pages={1067--1074},
  year={2022},
  publisher={American Association for the Advancement of Science}
}

@article{xu2023language,
  title={Language agents with reinforcement learning for strategic play in the werewolf game},
  author={Xu, Zelai and Yu, Chao and Fang, Fei and Wang, Yu and Wu, Yi},
  journal={arXiv preprint arXiv:2310.18940},
  year={2023}
}

@article{wu2024enhance,
  title={Enhance reasoning for large language models in the game werewolf},
  author={Wu, Shuang and Zhu, Liwen and Yang, Tao and Xu, Shiwei and Fu, Qiang and Wei, Yang and Fu, Haobo},
  journal={arXiv preprint arXiv:2402.02330},
  year={2024}
}

@article{guo2023suspicion,
  title={Suspicion-agent: Playing imperfect information games with theory of mind aware gpt-4},
  author={Guo, Jiaxian and Yang, Bo and Yoo, Paul and Lin, Bill Yuchen and Iwasawa, Yusuke and Matsuo, Yutaka},
  journal={arXiv preprint arXiv:2309.17277},
  year={2023}
}

@article{mao2024alympics,
  title={ALYMPICS: LLM Agents Meet Game Theory--Exploring Strategic Decision-Making with AI Agents},
  author={Mao, Shaoguang and Cai, Yuzhe and Xia, Yan and Wu, Wenshan and Wang, Xun and Wang, Fengyi and Ge, Tao and Wei, Furu},
  journal={arXiv preprint arXiv:2311.03220},
  year={2024}
}

@article{vickrey1961counterspeculation,
  title={Counterspeculation, auctions, and competitive sealed tenders},
  author={Vickrey, William},
  journal={The Journal of finance},
  volume={16},
  number={1},
  pages={8--37},
  year={1961},
  publisher={JSTOR}
}

@article{ledoux1981concours,
  title={Concours r{\'e}sultats complets},
  author={Ledoux, Alain},
  journal={Les victimes se sont plu {\`a} jouer le},
  volume={14},
  pages={10--11},
  year={1981}
}

@article{shinn2023reflexion,
  title={Reflexion: Language agents with verbal reinforcement learning},
  author={Shinn, Noah and Cassano, Federico and Gopinath, Ashwin and Narasimhan, Karthik and Yao, Shunyu},
  journal={Advances in Neural Information Processing Systems},
  volume={36},
  pages={8634--8652},
  year={2023}
}

@article{wei2022chain,
  title={Chain-of-thought prompting elicits reasoning in large language models},
  author={Wei, Jason and Wang, Xuezhi and Schuurmans, Dale and Bosma, Maarten and Xia, Fei and Chi, Ed and Le, Quoc V and Zhou, Denny and others},
  journal={Advances in neural information processing systems},
  volume={35},
  pages={24824--24837},
  year={2022}
}

@article{xu2023exploring,
  title={Exploring large language models for communication games: An empirical study on werewolf},
  author={Xu, Yuzhuang and Wang, Shuo and Li, Peng and Luo, Fuwen and Wang, Xiaolong and Liu, Weidong and Liu, Yang},
  journal={arXiv preprint arXiv:2309.04658},
  year={2023}
}

@article{guan2024richelieu,
  title={Richelieu: Self-evolving llm-based agents for ai diplomacy},
  author={Guan, Zhenyu and Kong, Xiangyu and Zhong, Fangwei and Wang, Yizhou},
  journal={Advances in Neural Information Processing Systems},
  volume={37},
  pages={123471--123497},
  year={2024}
}

@article{yang2020overview,
  title={An overview of multi-agent reinforcement learning from game theoretical perspective},
  author={Yang, Yaodong and Wang, Jun},
  journal={arXiv preprint arXiv:2011.00583},
  year={2020}
}

@article{Wang2022SelfConsistencyIC,
  author    = {Wang, Xuezhi and Wei, Jason and Schuurmans, Dale and Le, Quoc and Chi, Ed and Narang, Sharan and Chowdhery, Aakanksha and Zhou, Denny},
  title     = {Self-Consistency Improves Chain of Thought Reasoning in Language Models},
  journal   = {arXiv preprint arXiv:2203.11171},
  year      = {2022},
  url       = {https://arxiv.org/abs/2203.11171}
}

@article{yao2023tree,
  title={Tree of thoughts: Deliberate problem solving with large language models},
  author={Yao, Shunyu and Yu, Dian and Zhao, Jeffrey and Shafran, Izhak and Griffiths, Tom and Cao, Yuan and Narasimhan, Karthik},
  journal={Advances in neural information processing systems},
  volume={36},
  pages={11809--11822},
  year={2023}
}

@inproceedings{Besta2024GraphofThoughtsSE,
  title={Graph of thoughts: Solving elaborate problems with large language models},
  author={Besta, Maciej and Blach, Nils and Kubicek, Ales and Gerstenberger, Robert and Podstawski, Michal and Gianinazzi, Lukas and Gajda, Joanna and Lehmann, Tomasz and Niewiadomski, Hubert and Nyczyk, Piotr and others},
  booktitle={Proceedings of the AAAI conference on artificial intelligence},
  volume={38},
  number={16},
  pages={17682--17690},
  year={2024}
}

@article{Zhang2024OnTD,
  author    = {Zhang, Yilun and Cai, Yujun and Li, Yifei and Yang, Yaodong},
  title     = {On the Diagram of Thought},
  journal   = {arXiv preprint arXiv:2409.10038},
  year      = {2024},
  url       = {https://arxiv.org/abs/2409.10038}
}

@article{Li2025LogicofThoughtEL,
  author    = {Li, Naiqi and Liu, Peiyuan and Liu, Zheng and Dai, Tao and Jiang, Yong and Xia, Shu-Tao},
  title     = {Logic-of-Thought: Empowering Large Language Models with Logic Programs for Solving Puzzles in Natural Language},
  journal   = {arXiv preprint arXiv:2505.16114},
  year      = {2025},
  url       = {https://arxiv.org/abs/2505.16114}
}

@article{papoudakis2021agent,
title={Agent modelling under partial observability for deep reinforcement learning},
author={Papoudakis, Georgios and Christianos, Filippos and Albrecht, Stefano},
journal={Advances in Neural Information Processing Systems},
volume={34},
pages={19210--19222},
year={2021}
}

@article{zintgraf2021deep,
title={Deep interactive bayesian reinforcement learning via meta-learning},
author={Zintgraf, Luisa and Devlin, Sam and Ciosek, Kamil and Whiteson, Shimon and Hofmann, Katja},
journal={arXiv preprint arXiv:2101.03864},
year={2021}
}

@inproceedings{fu2022greedy,
title={Greedy when sure and conservative when uncertain about the opponents},
author={Fu, Haobo and Tian, Ye and Yu, Hongxiang and Liu, Weiming and Wu, Shuang and Xiong, Jiechao and Wen, Ying and Li, Kai and Xing, Junliang and Fu, Qiang and others},
booktitle={International Conference on Machine Learning},
pages={6829--6848},
year={2022},
organization={PMLR}
}

@inproceedings{kim2021policy,
title={A policy gradient algorithm for learning to learn in multiagent reinforcement learning},
author={Kim, Dong Ki and Liu, Miao and Riemer, Matthew D and Sun, Chuangchuang and Abdulhai, Marwa and Habibi, Golnaz and Lopez-Cot, Sebastian and Tesauro, Gerald and How, Jonathan},
booktitle={International Conference on Machine Learning},
pages={5541--5550},
year={2021},
organization={PMLR}
}

@article{yu2022model,
title={Model-based opponent modeling},
author={Yu, Xiaopeng and Jiang, Jiechuan and Zhang, Wanpeng and Jiang, Haobin and Lu, Zongqing},
journal={Advances in Neural Information Processing Systems},
volume={35},
pages={28208--28221},
year={2022}
}

@inproceedings{jing2025open,
title={An Open-Ended Learning Framework for Opponent Modeling},
author={Jing, Yuheng and Li, Kai and Liu, Bingyun and Fu, Haobo and Fu, Qiang and Xing, Junliang and Cheng, Jian},
booktitle={Proceedings of the AAAI Conference on Artificial Intelligence},
volume={39},
number={22},
pages={23222--23230},
year={2026}
}

@inproceedings{wu2022l2e,
title={L2E: Learning to exploit your opponent},
author={Wu, Zhe and Li, Kai and Xu, Hang and Zang, Yifan and An, Bo and Xing, Junliang},
booktitle={2022 International Joint Conference on Neural Networks (IJCNN)},
pages={1--8},
year={2022},
organization={IEEE}
}

@inproceedings{lu2022model,
title={Model-free opponent shaping},
author={Lu, Christopher and Willi, Timon and De Witt, Christian A Schroeder and Foerster, Jakob},
booktitle={International Conference on Machine Learning},
pages={14398--14411},
year={2022},
organization={PMLR}
}

@article{hu2023modeling,
title={Modeling opponent learning in multiagent repeated games},
author={Hu, Yudong and Han, Congying and Li, Haoran and Guo, Tiande},
journal={Applied Intelligence},
volume={53},
number={13},
pages={17194--17210},
year={2023},
publisher={Springer}
}

@inproceedings{zhang2024proagent,
  title={Proagent: building proactive cooperative agents with large language models},
  author={Zhang, Ceyao and Yang, Kaijie and Hu, Siyi and Wang, Zihao and Li, Guanghe and Sun, Yihang and Zhang, Cheng and Zhang, Zhaowei and Liu, Anji and Zhu, Song-Chun and others},
  booktitle={Proceedings of the AAAI Conference on Artificial Intelligence},
  volume={38},
  number={16},
  pages={17591--17599},
  year={2024}
}

@article{ji2023survey,
  title={Survey of hallucination in natural language generation},
  author={Ji, Ziwei and Lee, Nayeon and Frieske, Rita and Yu, Tiezheng and Su, Dan and Xu, Yan and Ishii, Etsuko and Bang, Ye Jin and Madotto, Andrea and Fung, Pascale},
  journal={ACM computing surveys},
  volume={55},
  number={12},
  pages={1--38},
  year={2023},
  publisher={ACM New York, NY}
}

@article{liu2023lost,
  title={Lost in the middle: How language models use long contexts},
  author={Liu, Nelson F and Lin, Kevin and Hewitt, John and Paranjape, Ashwin and Bevilacqua, Michele and Petroni, Fabio and Liang, Percy},
  journal={arXiv preprint arXiv:2307.03172},
  year={2023}
}

@book{pearl2000causality, 
  title={Causality: Models, Reasoning, and Inference},  
  author={Pearl, J.},
  year={2000},
  publisher={Cambridge University Press},  
  address={New York, NY, USA} 
}

@article{zhang2024k,
  title={K-Level Reasoning: Establishing Higher Order Beliefs in Large Language Models for Strategic Reasoning},
  author={Zhang, Yadong and Mao, Shaoguang and Ge, Tao and Wang, Xun and Xia, Yan and Lan, Man and Wei, Furu},
  journal={arXiv preprint arXiv:2402.01521},
  year={2024}
}

@article{xu2023magic,
  title={Magic: Investigation of large language model powered multi-agent in cognition, adaptability, rationality and collaboration},
  author={Xu, Lin and Hu, Zhiyuan and Zhou, Daquan and Ren, Hongyu and Dong, Zhen and Keutzer, Kurt and Ng, See Kiong and Feng, Jiashi},
  journal={arXiv preprint arXiv:2311.08562},
  year={2023}
}

@inproceedings{yang2025uncertainty,
  title={Uncertainty-Aware Opponent Modeling for Deep Reinforcement Learning},
  author={Yang, Likun and Xu, Pei and Cao, Shiyue and Ren, Yongjian and Chen, Xiaotang and Huang, Kaiqi},
  booktitle={Proceedings of the 24th International Conference on Autonomous Agents and Multiagent Systems},
  pages={2217--2225},
  year={2025}
}

\clearpage
\includepdf[pages=-,pagecommand={\thispagestyle{empty}}]{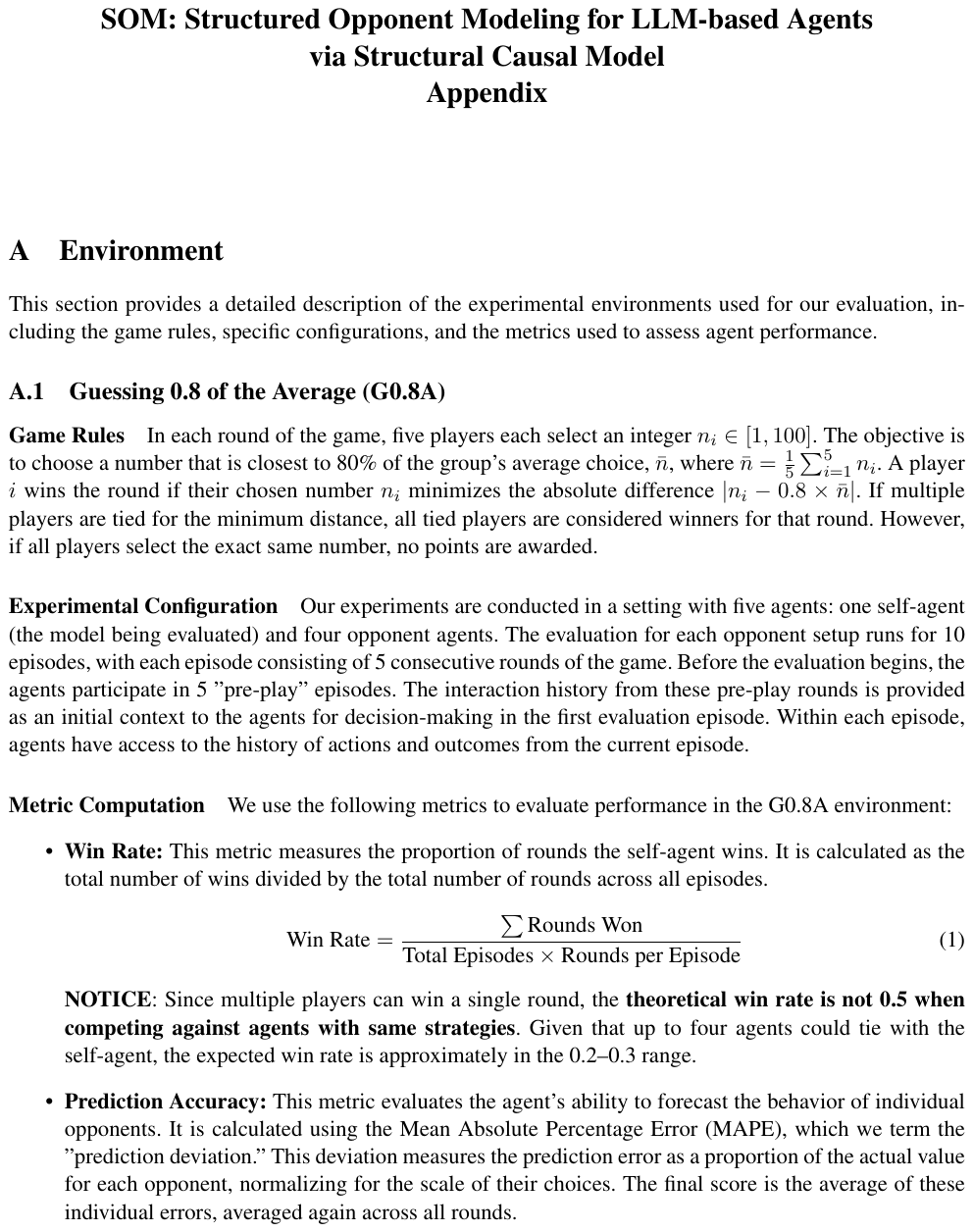}

\end{document}